\documentclass[runningheads]{template/llncs}

\usepackage{booktabs}
\usepackage{graphicx}
\usepackage{wrapfig}

\newcommand{\repeatthanks}{\textsuperscript{\thefootnote}}
\usepackage{xcolor}
\renewcommand{\labelenumi}{(\theenumi)}

\usepackage{fancyhdr}
\begin{document}
\renewcommand{\labelenumi}{\alph{enumi})}
\title{
    Sentiment Analysis with \\
    Contextual Embeddings and Self-Attention
}
\titlerunning{Sentiment Analysis with Contextual Embeddings \& Self-Attention}

\author{
    Katarzyna Biesialska
    \thanks{Both authors contributed equally to this work, which was mostly done at the Warsaw University of Technology.}
    \inst{1}\orcidID{0000-0002-2865-7990} 
    \and \\
    Magdalena Biesialska\repeatthanks\inst{1}\orcidID{0000-0001-7890-3523} 
    \and \\
    Henryk Rybinski\inst{2}\orcidID{0000-0002-2890-7080}
}
\authorrunning{K. Biesialska, M. Biesialska \& H. Rybinski}
\institute{Universitat Polit\`ecnica de Catalunya, Barcelona, Spain \\
\email{\{katarzyna,magdalena\}.biesialska@upc.edu} \and
Warsaw University of Technology, Warsaw, Poland
\email{h.rybinski@ii.pw.edu.pl}}
\maketitle              
%
\begin{abstract}
In natural language the intended meaning of a word or phrase is often implicit and depends on the context. In this work, we propose a simple yet effective method for sentiment analysis using contextual embeddings and a self-attention mechanism. The experimental results for three languages, including morphologically rich Polish and German, show that our model is comparable to or even outperforms state-of-the-art models. In all cases the superiority of models leveraging contextual embeddings is demonstrated. Finally, this work is intended as a step towards introducing a universal, multilingual sentiment classifier.

\end{abstract}
\section{Introduction}
All areas of human life are affected by people's views. With the sheer amount of reviews and other opinions over the Internet, there is a need for automating the process of extracting relevant information. For machines, however,  measuring sentiment is not an easy task, because natural language is highly ambiguous at all levels, and thus difficult to process. For instance, a single word can hardly convey the whole meaning of a statement. Moreover, computers often do not distinguish literal from figurative meaning or incorrectly handle complex linguistic phenomena, such as: sarcasm, humor, negation etc.

In this paper, we take a closer look at two factors that make automatic opinion mining difficult -- the problem of representing text information, and
sentiment analysis (SA). In particular, we leverage contextual embeddings, which enable to convey a word meaning depending on the context it occurs in. Furthermore, we build a hierarchical multi-layer classifier model, based on an architecture of the Transformer encoder \cite{vaswani2017attention}, 
primarily relying on a self-attention mechanism and bi-attention. The proposed sentiment classification model is language independent, which is especially useful for low-resource languages (e.g. Polish).




We evaluate our methods on various standard datasets, which allows us to compare our approach against current state-of-the-art models for 
three languages: English, Polish and German.
We show that our approach is comparable to the best performing sentiment classification models; and, importantly, in two cases yields significant improvements over the state of the art.

The paper is organized as follows: Section 2 presents the background and related work. Section 3 describes our proposed method. Section 4 discusses datasets, experimental setup, and results. Section 5 concludes this paper and outlines the future work.

\section{Related Work}

Sentiment classification has been one of the most active research areas in natural language processing (NLP) and has become one of the most popular downstream tasks to evaluate performance of neural network (NN) based models. The task itself encompasses several different opinion related tasks, hence it tackles many challenging NLP problems, see e.g. \cite{Liu2012SentimentAA,Mohammad2016SentimentAD}.

\subsection{Sentiment Analysis Approaches}
The first fully-formed techniques for SA emerged around two decades ago, and continued to be prevalent for several years, until deep learning methods entered the stage.
The most straight-forward method, developed in \cite{turney-2002-thumbs}, is based on the number of positive and negative words in a piece of text. Concretely, the text is assumed to have positive polarity if it contains more positive than negative terms, and vice versa. Of course, the term-counting method is often insufficient; therefore, an improved method was proposed in \cite{Kennedy2006SentimentCO}, which combines counting positive and negative terms with a machine learning (ML) approach (i.e. Support Vector Machine). 

Various studies (e.g. \cite{Turney2010FromFT}) have shown that one can determine the polarity of an unknown word by calculating co-occurrence statistics of it. 
Moreover, classical solutions to the SA problem are often based on lexicons. Traditional lexicon-based SA leverages word-lists, that are pre-annotated with positive and negative sentiment. Therefore, for many years lexicon-based approaches have been utilized when there was insufficient amount of labeled data to train a classifier in a fully supervised way. 

In general, ML algorithms are popular methods for determining sentiment polarity. 
A first ML model applied to SA has been implemented in \cite{pang2002thumbs}. Moreover, throughout the years, different variants of NN architectures have been introduced in the field of SA. Especially recursive neural networks \cite{Paulus2014GlobalBR}, such as recurrent neural networks (RNN) \cite{Socher2013RecursiveDM,tai-etal-2015-improved,Kumar2015AskMA}, or convolutional neural networks (CNN) \cite{Kalchbrenner2014ACN,Kim2014ConvolutionalNN} have become the most prevalent choices. 


\subsection{Vector Representations of Words}
One of the principal concepts in linguistics states that related words can be used in similar ways \cite{firth1957synopsis}. Importantly, words may have different meaning in different contexts. Nevertheless, until recently it has been a dominant approach (e.g. word2vec \cite{Mikolov2013DistributedRO}, GloVe \cite{Pennington2014GloveGV}) to learn representations such that each and every word has to capture all its possible meanings. 

However, lately a new set of methods to learn dynamic representations of words has emerged \cite{mccann2017learned,howard2018universal,Peters2018,radfordimproving,devlin2019bert}. These approaches allow each word representation to capture what a word means in a particular context. While every word token has its own vector, the vector can depend on a variable-length sequence of nearby words (i.e. context). Consequently, a context vector is obtained by feeding a neural network with these context word vectors and subsequently encoding them into a single fixed-length vector.
\begin{wrapfigure}{R}{0.5\textwidth}
\centering
\includegraphics[width=0.5\textwidth]{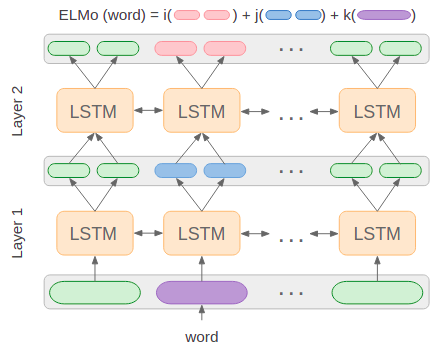}
\caption{\label{fig:elmo}The architecture of ELMo.}
\end{wrapfigure}

ULMFiT \cite{howard2018universal} was the very first method to induce contextual word representations by harnessing the power of language modeling. The authors proposed to learn contextual embeddings by pre-training a language model (LM), and then performing task-specific fine-tuning. ULMFiT architecture is based on a vanilla 3-layer Long Short-Term Memory (LSTM) NN without any attention mechanism. 

The other contextual embedding model introduced recently is called ELMo (Embeddings from Language Models) \cite{Peters2018}. Similarly to ULMFiT, this model uses tokens at the word-level. ELMo contextual embeddings are “deep” as they are a function of all hidden states. Concretely, context-sensitive features are extracted from a left-to-right and a right-to-left 2-layer bidirectional LSTM language models. Thus, the contextual representation of each word is the concatenation of the left-to-right and right-to-left representations as well as the initial embedding (see Fig. \ref{fig:elmo}). 


The most recent model -- BERT \cite{devlin2019bert} -- is more sophisticated architecturally-wise, as it is a multi-layer masked LM based on the Transformer NN utilizing sub-word tokens. However, as we are bound to use word-level tokens in our sentiment classifier, we leverage the ELMo model for obtaining contextual embeddings. More specifically, by means of ELMo we are able to feed our classifier model with context-aware embeddings of an input sequence. Hence, in this setting we do not perform any fine-tuning of ELMo on a downstream task. 

\subsection{Self-Attention Deep Neural Networks}
The attention mechanism was introduced in
\cite{bahdanau2014neural} in 2014 
and since then it has been applied  successfully to different computer vision (e.g. visual explanation) 
and NLP (e.g. machine translation) tasks. The mechanism is often used as an extra source of information added on top of the CNN or LSTM model to enhance the extraction of sentence embedding \cite{Santos2016AttentivePN,Lin2017ASS}. However,
this scenario is not applicable to sentiment classification, since the model only receives a single sentence on input, hence there is no such extra information  \cite{Lin2017ASS}.


Self-attention (or intra-attention) is an attention mechanism that computes a representation of a sequence by relating different positions of a single sequence. 
Previous work on sentiment classification has not covered extensively
attention-based neural network models for SA (especially using the Transformer architecture \cite{vaswani2017attention}), although some papers 
have appeared recently \cite{ambartsoumian2018self,letarte2018importance}. 

\section{The Proposed Approach}

\begin{wrapfigure}{R}{0.35\textwidth}
\centering
\includegraphics[width=0.27\textwidth]{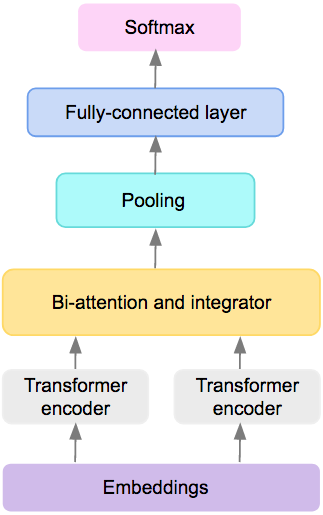}
\caption{\label{fig:tsa}An overview of the TSA model architecture.}
\end{wrapfigure}

Our proposed model, called Transformer-based Sentiment Analysis (TSA) (see Fig. \ref{fig:tsa}), is based on the recently introduced Transformer architecture \cite{vaswani2017attention}, which has provided significant improvements for the neural machine translation task. 
Unlike RNN or CNN based models, the Transformer is able to learn dependencies between distant positions. Therefore, in this paper we show that attention-based models are suitable for other NLP tasks, such as learning distributed representations and sentiment analysis, and thus are able to improve the overall accuracy.

The architecture of the TSA model and steps to train it can be summarized as follows:

\renewcommand{\theenumi}{\roman{enumi}}%
\begin{enumerate}
\item At the very beginning there is a simple text pre-processing method that performs text clean-up and splits text into tokens.
\item We use contextual word representations to represent text as real-valued vectors.
\item After embedding the text into real-valued vectors, the Transformer network maps the input sequence into hidden states using self-attention.
\item Next a bi-attention mechanism is utilized to estimate the interdependency between representations.
\item A single layer LSTM together with self-attentive pooling compute the pooled representations.
\item A joint representation for the inputs is later passed to a fully-connected neural network.
\item Finally, a softmax layer is used to determine sentiment of the text.

\end{enumerate}

\subsection{Embeddings and Encoded Positional Information}

Non-recurrent models, such as deep self-attention NN, do not necessarily process the input sequence in a sequential manner. Hence, there is no way they can record the position of each word in a sequence, which is an inherent limitation of every such model. Therefore, in the case of the Transformer, the need has been addressed in the following manner -- the Transformer takes into account the order of the words in the input sequence by encoding their position information in extra vectors (so called positional encoding vectors) and adding them to input embeddings. There are many different approaches to embed position information, such as learned or fixed positional encodings (PE), or recently introduced relative position representations (RPR) \cite{shaw2018}.
The original Transformer used sine and cosine functions of different frequencies.

In this work, we explore the effectiveness of applying a modified approach to incorporate positional information into the model, namely using RPR instead of 
PE. Furthermore, we use global average pooling in order to average the output of the last self-attention layer and prepare the model for the final classification layer.

\subsection{The Transformer Encoder}
The input sequence is combined with word and positional embeddings, which provide time signal, and together are fed into an encoder block. Matrices for a query \textit{Q}, a key \textit{K} and a value \textit{V} are calculated and passed to a self-attention layer. Next, a normalization is applied and residual connections provide additional context. Further, a final dense layer with vocabulary size generates the output of the encoder. A fully-connected feed-forward network within the model is a single hidden layer network with a ReLU activation function in between:
\begin{equation}
FFN(x)=\max\left(0, x W_{1}+b_{1}\right) W_{2}+b_{2}
\end{equation}

\subsection{Self-Attention Layer}
The self-attention block in the encoder is called multi-head self-attention. A self-attention layer allows each position in the encoder to access all positions in the previous layer of the encoder immediately, and in the first layer all positions in the input sequence. The multi-head self-attention layer employs \textit{h} parallel self-attention layers, called heads, with different \textit{Q}, \textit{K}, \textit{V} matrices obtained for each head. In a nutshell, the attention mechanism in the Transformer architecture relies on a scaled dot-product attention, which is a function of \textit{Q} and a set of \textit{K}-\textit{V} pairs. The computation of attention is performed in the following order. First, a multiplication of a query and transposed key is scaled through the scaling factor of $1/\sqrt{d_{z}}$ (Eq. 2)
\begin{equation}
m_{i j}=\frac{Q K^{T}}{\sqrt{d_{z}}}
\end{equation}
Next, the attention is produced using the softmax function over their scaled inner product:
\begin{equation}
\alpha_{i j}=\frac{e^{m_{i j}}}{\sum_{k=1}^{n} e^{m_{i k}}}
\end{equation}
Finally, the weighted sum of each attention head and a value is calculated as follows:
\begin{equation}
z_{i}=\sum_{j=1}^{n} \alpha_{i j} V
\end{equation}

\subsection{Masking and Pooling}
Similar to other sources of data, the datasets used for training and evaluation of our models contain sequences of different length. The most common approach in the literature involves finding a maximal sequence length existing in the dataset/batch and padding sentences that are shorter than the longest one with trailing zeroes.
In the proposed TSA model, we deal with the problem of variable-length sequences by using masking and self-attentive pooling. The inspiration for our approach comes from the BCN model proposed in \cite{mccann2017learned}. Thanks to this mechanism, we are able to fit sequences of different length into the final fixed-size vector, which is required for the computation of the sentiment score. The self-attentive pooling layer is applied just after the encoder block. 

\section{Experiments}

\subsection{Datasets}
In this work, we compare sentiment analysis results considering four benchmark datasets in three languages. All datasets are originally split into training, dev and test sets. Below we describe these datasets in more detail.

\begin{table}
\centering
\small
\caption{Sentiment analysis datasets with
number of classes and train/dev/test split.}\label{tab1}
\setlength{\tabcolsep}{0.5em}
\begin{tabular}{|l|c|r|r|r|c|c|}
\hline
\textbf{Dataset} & \textbf{\# Classes} & \textbf{Train} & \textbf{Dev} & \textbf{Test} & \textbf{Domain} & \textbf{Language} \\
\hline
SST-2 & 2 & 6,920 & 872 & 1,821 & movies & English \\
SST-5 & 5 & 8,544 & 1,101 & 2,210 & movies & English \\
PolEmo 2.0-IN & 5 & 5,783 & 723 & 722 & medical, hotels & Polish \\
GermEval & 3 & 19,432 & 2,369 & 2,566 & travel, transport & German \\
\hline
\end{tabular}
\end{table}

\paragraph{\textbf{Stanford Sentiment Treebank (SST)}} This collection of movie reviews \cite{Socher2013RecursiveDM} from the
\texttt{rottentomatoes.com} is annotated for the binary (SST-2) and fine-grained (SST-5) sentiment classification. SST-2 divides reviews into two groups: \textit{positive} and \textit{negative}, while SST-5 distinguishes 5 different review types: \textit{very positive},\textit{ positive}, \textit{neutral}, \textit{negative}, \textit{very negative}. The dataset consists of 11,855 single sentences and is widely used in the NLP community.
\paragraph{\textbf{PolEmo 2.0}} The dataset \cite{kocon-etal-2019-multi} comprises online reviews from education, medicine and hotel domains. There are two separate test sets, to allow for in-domain (medicine and hotels) and out-of-domain (products and university) evaluation. The dataset comes with the following sentiment labels: \textit{strong positive}, \textit{weak positive}, \textit{neutral}, \textit{weak negative}, \textit{strong negative}, and \textit{ambiguous}.
\paragraph{\textbf{GermEval}} This dataset \cite{germevaltask2017} contains customer reviews of the railway operator (Deutsche Bahn) published on social media and various web pages. Customers expressed their feedback regarding the service of the railway company (e.g. travel experience, timetables, etc.) by rating it as \textit{positive}, \textit{negative}, or \textit{neutral}.

\subsection{Experimental Setup}
Pre-processing of input datasets is kept to a minimum as we perform only tokenization when required. Furthermore, even though some datasets, such as SST or GermEval, provide additional information (i.e. phrase, word or aspect-level annotations), for each review we only extract text of the review and its corresponding rating.

The model is implemented in the Python programming language, PyTorch\footnote{https://pytorch.org} and AllenNLP\footnote{https://allennlp.org}. Moreover, we use pre-trained word-embeddings, such as ELMo \cite{Peters2018}, GloVe \cite{Pennington2014GloveGV}. Specifically, we use the following ELMo models: Original\footnote{https://allennlp.org/elmo}, Polish \cite{janz-elmo-2019} and German \cite{GerElmo}. In the ELMO+GloVe+BCN model we use the following 300-dimension GloVe embeddings: English\footnote{http:$//$nlp.stanford.edu$/$data$/$glove.840B.300d.zip}, Polish \cite{dadas2019} and German\footnote{https://wikipedia2vec.github.io/wikipedia2vec/pretrained}. In order to simplify our approach when training the sentiment classifier model, we establish a very similar setting to the vanilla Transformer. We use the same optimizer - Adam with $\beta_{1}=0.9, \beta_{2}=0.98$, and $\epsilon=10^{-9}$. We incorporate four types of regularization during training: dropout probability $P_{drop}=0.1$, embedding dropout probability $P_{emb}=0.5$, residual dropout probability $P_{res}=0.2$, and attention dropout probability $P_{attn}=0.1$. We use 2 encoder layers. In addition, we employ label smoothing of value $\epsilon_{ls}=0.1$. In terms of RPR parameters, we set clipping distance to $k=10$.


\subsection{Results and Discussion}

In Table~\ref{tab:results}, we summarize experimental results achieved by our model and other state-of-the-art systems reported in the literature by their respective authors.

\begin{table}
\centering
\small
\caption{Results of our systems compared to baselines and state-of-the-art systems evaluated on English, Polish and German sentiment classification datasets.}
\label{tab:results}
\setlength{\tabcolsep}{0.525em}
{%
\begin{tabular}{lcccc}
\toprule
 & \multicolumn{2}{c}{\textbf{English}} & \multicolumn{1}{c}{\textbf{Polish}} & \multicolumn{1}{c}{\textbf{German}} \\ \cmidrule{2-5} 
 & SST-2 & SST-5 & PolEmo2.0-IN & GermEval \\ \hline
RNTN \cite{Socher2013RecursiveDM} & 85.4 & 45.7 & - & - \\
DCNN \cite{Kalchbrenner2014ACN} & 86.8 & 48.5 & - & - \\
CNN \cite{Kim2014ConvolutionalNN} & 88.1 & 48.0  & - & - \\
DMN \cite{Kumar2015AskMA} & 88.6 & 52.1 & - & - \\
Constituency Tree-LSTM \cite{tai-etal-2015-improved} & 88.0 & 51.0 & - & - \\
CoVe+BCN \cite{mccann2017learned} & 90.3 & \textbf{53.7} & - & - \\
SSAN+RPR \cite{ambartsoumian2018self} & 84.2 & 48.1 & - & - \\
Polish BERT \cite{KLEJ} & - & - & 88.1 & - \\
SWN2-RNN \cite{germevaltask2017} & - & - & - & 74.9 \\
\midrule 
& \multicolumn{2}{c}{\textit{Our baseline}} \\
\midrule
ELMo+GloVe+BCN & \textbf{91.4} & 53.5 & 88.9 & 78.2 \\
\midrule 
& \multicolumn{2}{c}{\textit{Our model}} \\
\midrule
ELMo+TSA & \multicolumn{1}{c}{89.3} & \multicolumn{1}{c}{50.6} &  \multicolumn{1}{c}{\textbf{89.8}} & \textbf{78.9} \\
\bottomrule
\end{tabular}%
}
\end{table}

We observe that our models, baseline and ELMo+TSA, outperform state-of-the-art systems for all three languages. More importantly, the presented accuracy scores indicate that the TSA model is competitive and for two languages (Polish and German) achieves the best results. Also noteworthy, in Table~\ref{tab:results}, there are two models that use some variant of the Transformer: SSAN+RPR \cite{ambartsoumian2018self} uses the Transformer encoder for the classifier, while Polish BERT \cite{KLEJ} employs Transformer-based language model introduced in \cite{devlin2019bert}. One of the reasons why we achieve higher score for the SST dataset might be that the authors of SSAN+RPR used word2vec embeddings \cite{Mikolov2013DistributedRO}, whereas we employ ELMo contextual embeddings \cite{Peters2018}. Moreover, in our TSA model we use not only self-attention (as in SSAN+RPR) but also a bi-attention mechanism, hence this also should provide performance gains over standard architectures. 

In conclusion, comparing the results of the models leveraging contextual embeddings (CoVe+BCN, Polish BERT, ELMo+GloVe+BCN and ELMo+TSA) with the rest of the reported models, which use traditional distributional word vectors, we note that the former category of sentiment classification systems demonstrates remarkably better results.

\section{Conclusion and Future Work}
We have presented a novel architecture, based on the Transformer encoder with relative position representations.  Unlike existing models, this work proposes a model relying solely on a self-attention mechanism and bi-attention. We
show that our sentiment classifier model achieves very good results, comparable to the state of the art, even though it is language-agnostic. Hence, this work is a step towards building a universal, multi-lingual sentiment classifier. 

In the future, we plan to evaluate our model using benchmarks also for other languages. It is particularly interesting to analyze the behavior of our model with respect to low-resource languages. Finally, other promising research avenues worth exploring are related to unsupervised cross-lingual sentiment analysis.

\bibliographystyle{template/splncs04}
\bibliography{bibliography}

\end{document}